# MODELLING REACTIVE AND PROACTIVE BEHAVIOUR IN SIMULATION

*Mrs Mazlina Abdul Majid*
*Dr Peer-Olaf Siebers*
*Professor Uwe Aickelin*

School of Computer Science
University of Nottingham,
Nottingham, UK
mva@cs.nott.ac.uk, pos@cs.nott. ac.uk, uxa@cs.nott.ac.uk

**ABSTRACT:**
*This research investigated the simulation model behaviour of a traditional and combined discrete event as well as agent based simulation models when modelling human reactive and proactive behaviour in human centric complex systems. A departmental store was chosen as human centric complex case study where the operation system of a fitting room in WomensWear department was investigated. We have looked at ways to determine the efficiency of new management policies for the fitting room operation through simulating the reactive and proactive behaviour of staff towards customers. Once development of the simulation models and their verification had been done, we carried out a validation experiment in the form of a sensitivity analysis. Subsequently, we executed a statistical analysis where the mixed reactive and proactive behaviour experimental results were compared with some reactive experimental results from previously published works. Generally, this case study discovered that simple proactive individual behaviour could be modelled in both simulation models. In addition, we found the traditional discrete event model performed similar in the simulation model output compared to the combined discrete event and agent based simulation when modelling similar human behaviour.*

## 1. INTRODUCTION

Simulation has become a preferred tool in Operation Research (OR) for modelling complex systems. Studies in human behaviour have increasingly attracted interest and attention from simulation research in the UK (Robinson, 2004). Discrete Event Simulation (DES) and Agent Based Simulation (ABS) are the two common simulation approaches applied for modelling human behaviour in OR. The capability of modelling human behaviour in both simulation approaches depends on their ability to model diverse and heterogeneous populations, while System Dynamics Simulation (SDS) which is another typical tool for system analysis is only appropriate to model up populations at an aggregate level. Nevertheless, the question is: What kind of human behaviour can be modelled with standard model designs in DES and ABS and how similar the results would be when one models the same human centric system with those two different approaches? When discussing different kinds of human behaviour, we refer to reactive and proactive human behaviour. On top of that, Rank et al. (2007) stressed that proactive behaviour is a very important aspect to succeed in the service industry. Therefore, we picked retailing as part of the service industry for our investigation in this study.

Undeniable, there are loads of human interactions in retailing. In this study, reactive behaviour refers to response of an available staff member to the customers' requests while proactive behaviour relates to a staff member's personal initiative in identifying and solving an issue. In retail, both behaviours; particularly proactive, play important roles in an organisation's ability to generate income and revenue. Therefore, our general aim was to investigate the usefulness of modelling human reactive and proactive behaviour in the retail performance using a combined DES/ABS approach. To achieve the aim, we produced a research question: What are the similarities and dissimilarities can be identified by modelling human reactive behaviour compared to both human reactive and proactive behaviour in the simulation output performance? In answering the research question, we carried out a case study in a departmental store which is one of the top ten retailers in the UK. In previous work, we had studied capabilities of traditional DES and combined DES/ABS in representing the impact of reactive staff behaviour on system performance (Majid et al, 2009). Currently, this paper focused on abilities of traditional DES and combined DES/ABS in representing the impact of mixed reactive and proactive staff behaviour on system performance. At the end of the study, findings of both studies were compared. Statistical tests will be used to establish if the dissimilarities in output performance of the different modelling methods are significant. From now on, we will refer the traditional DES model as the DES model whilst the combined DES/ABS model as the DES/ABS model.

The remaining content of this paper is presented as follows: Section 2 explores theory and characteristics of three major OR simulation methods - DES, ABS, and SDS. This section also discusses the comparisons between different simulation methods that were obtained from the literature. Then, Section 3 describes our case study plan, fieldwork and model design. Next, Section 4 presents our experimental setup, results, analysis as well as some discussion on the results. Finally, Section 5 will conclude the overall study as well as summarise our current progress.

## 2. BACKGROUND

### 2.1 Simulation

Over the last three decades, simulation studies have widely been used as a decision support tool in OR (Kelton, 2007). The ability to support studies of complex systems has made simulation as the most preferred choice among academicians and practitioners as compared to analytical tools. The simulation modelling paradigms used in OR can be classified in three groups: DES modelling, ABS modelling, and SDS modelling.

DES models represent systems based on chronological sequences of events (technically there is only one thread of execution, the system is centralised) where each event changes the system state in discrete time. It is difficult to represent proactive behaviour in DES models as usually people are implemented as resources or passive entities. Passive entities are unable to initiate events. Therefore, a proactive event that requires self-started behaviour by an individual entity is difficult to be implemented in DES.

ABS models (also called individual based models) however, offer a more straightforward solution to this issue. They contain active objects (agents) which possess the abilities of being autonomous, responsive, proactive and social (Jennings et al, 1998). ABS models are essentially decentralised, which means there is no place where the global system behaviour (dynamics) is defined. The modeller defines the behaviour at the micro level (individual level) and the macro behaviour (system behaviour) emerges from the many interactions between the individual entities (Borshchev and Filippov, 2004). As agents are self-controlled (technically, every agent has its own thread of execution, hence, the system is decentralised) they can initiate events independently. In addition, with ABS supports communication among the agents through message passing, which is a useful asset for modelling human behaviour.

SDS models represent a real world phenomenon using stock and flow diagrams, causal loop diagrams and differential equations. The inability of the SDS models to model heterogeneous individual behaviour is the main limitation of this approach in achieving the goal of this study. SDS models are unable to represent specific behaviour such as reactive or proactive behaviour of individuals in a system. Consequently, we excluded SDS modelling from our research study.

### 2.2 Retail and proactive behaviour modelling

The retail sector has been one of the main contributors to the global market economy. Excellent customer service is vital in products marketing as well as to maintain regular and to attract new customers. A good service is closely related to the ways staffs provides support to the customers. This behaviour is known as customer service which is a part of every company's management strategy. Service behaviour of a staff member can be classified as either reactive or proactive towards the customer.

According to Rank et al. (2007), since the 1990s various forms of proactive behaviour have been studied by various organisation scientists. This is because, implementing proactive behaviour provides potential success in career development, organisational change, stress management, etc. (Crant, 2000). On the other hand, to understand the potential benefit of supporting proactive behaviour of a staff member, we need to study their behavioural performance using OR methods. Siebers et al. (2009) has suggested the use of simulation when one is interested in the development of a system over time. The authors also emphasised that modelling and simulation of operational management practices in the retail sector are fairly common, while people management practices are often ignored. We found that most of these models only involve modelling of human reactive behaviour. Therefore, in order to understand the integration of operational management practices with human service behaviour, we need to come up with a research that involves modelling human reactive and proactive service behaviour. For such a study, we need to identify the appropriate tools.

**2.3 The Simulation choice**

This study was interested to investigate a service oriented system in the retail sector, which involved queuing for the different services. As in ABS models, the system itself is not explicitly modelled but emerges from the interaction of the many individual entities that make up the system, therefore, using ABS alone would not correspond to our investigations. However, as ABS seems to be a good concept for representing human behaviour, we decided to try out a combined DES/ABS approach where we modelled the system in a process-oriented manner while we modelled the actors inside the system (customers and staff) as agents. At the end of the study, we compared this approach with a more traditional DES approach.

**3. FIELDWORK**

Our case study focused on the operations in the main fitting room in a WomensWear (WW) department of a departmental store (see Figure 1). Simulation study is important for manager as it could help in identifying the potential impact on fitting room performance when having different numbers of fitting room cubicles, different numbers of fitting room staff, different staff roles, or empowering staff.

We simplify the real world reactive and proactive behaviour of the staff towards their customers and investigate how the behaviour effects on our simulation models. Reactive behaviour refers to staff's response to customer when they are available and requested. Typically, a staff in the fitting room has to do three reactive behaviour tasks: (1) counts the number of clothes and hands out a plastic card which contains the number of clothes taken in and the room number, (2) provides help while customers are in the fitting room, and (3) collects back the plastic card and any unwanted clothes when the customer leaves the fitting room area. On the other hand, proactive behaviour refers to a staff's self started behaviour, for example to deal with various demands. The proactive behaviour that we focused on was when a staff speeded up their service when the fitting room was getting hectic which resulted to time-consuming service and customers had to wait long to be served.

Based on our case study observations in the real system, we developed the concept for a DES model that represented the basic process flow of the department operations (a complex queuing system). We used this as a basis for both our conceptual models (for implementing the DES and the DES/ABS model later).

Then we developed concepts of the proactive behaviour representation for a process centric approach to be used in the DES model (see Figure 2) and for an individual centric approach to be used in the DES/ABS model (see Figure 3).

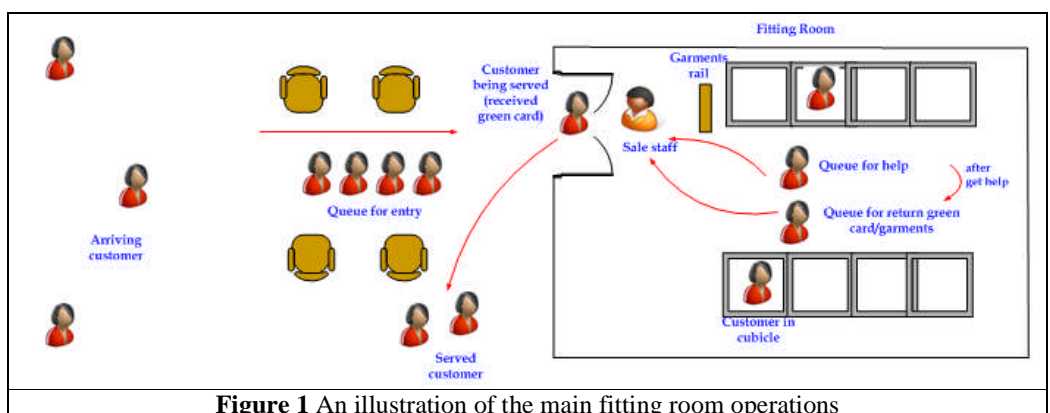

**Figure 1** An illustration of the main fitting room operations

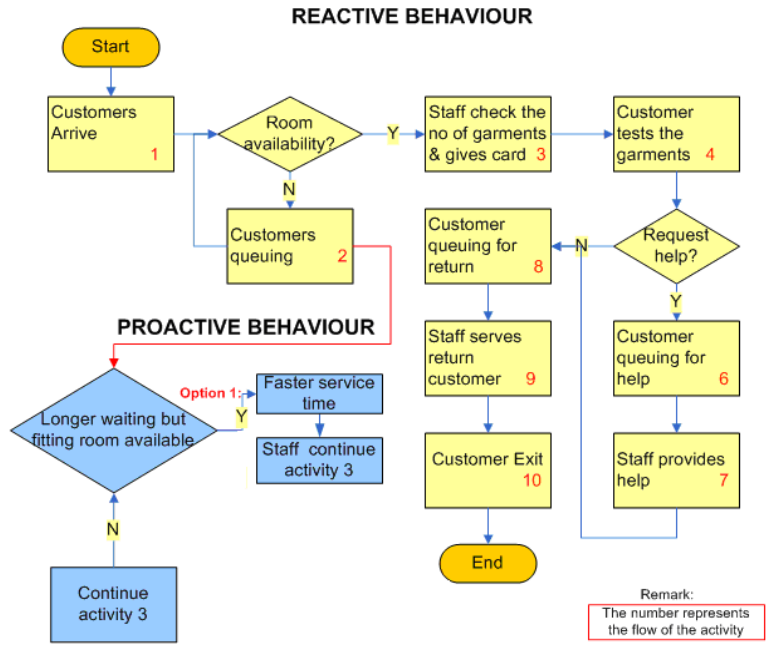

**Figure 2** Flow chart for DES

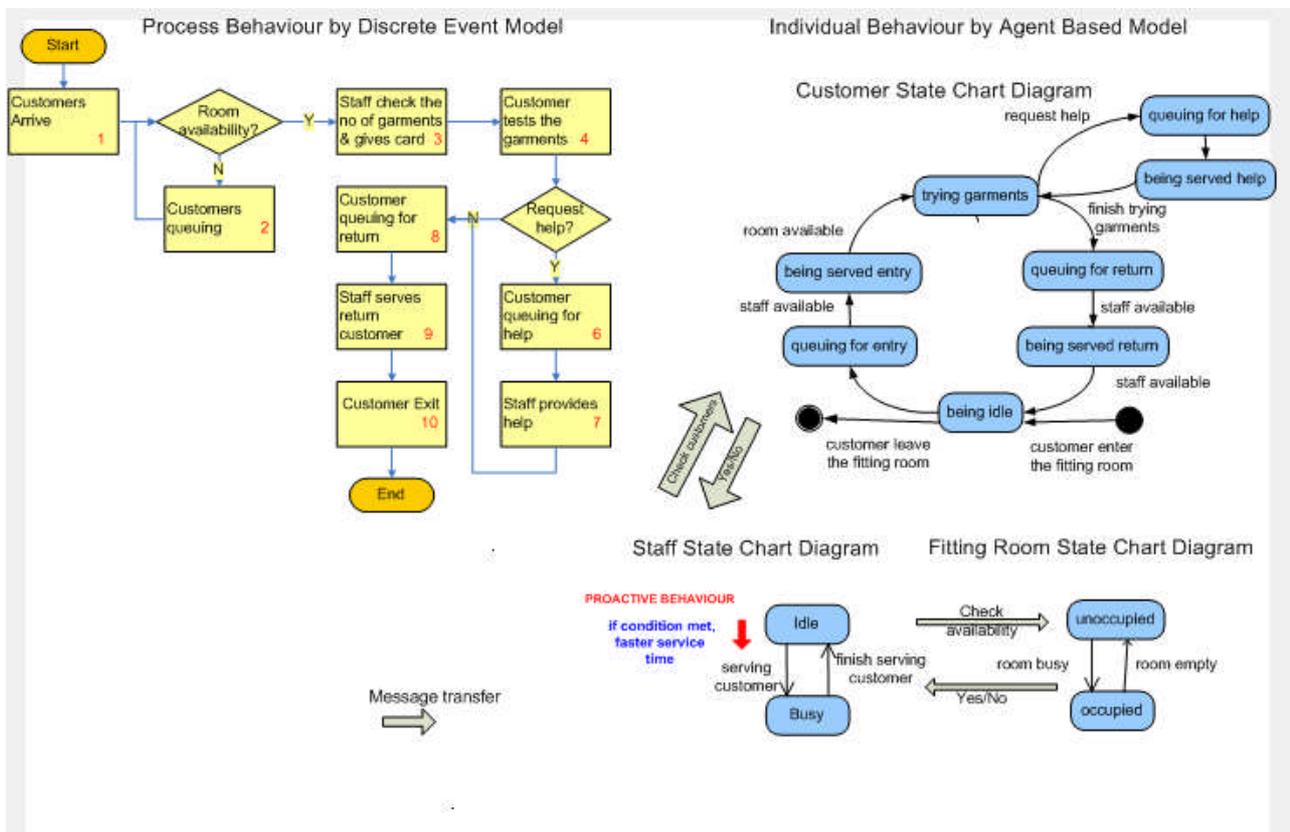

**Figure 3** Flow and state charts for DES/ABS

The figure shows that a single process flow chart diagram was used for the conceptual DES model. Meanwhile, for the conceptual DES/ABS model, we employed a process flow chart diagram and additional state charts diagrams to represent different types of agents (customers, staff, and fitting rooms). State charts show the different states an entity could be in and define the events that cause a transition from one state to another.

## 4. EXPERIMENTATION

### 4.1 Basic Model Setup

Based on both conceptual models presented in Section 3, we developed two simulation models. These models were implemented in the multi-paradigm simulation software AnyLogic™ (XJ Technologies, 2009). Both simulation models consisted of an arrival process (customers), three single queues (entry queue, return queue, help queue), and resources (one sales staff, one fitting room with 8 fitting cubicles). In our DES model customers, staff, and fitting rooms were all passive objects while in our DES/ABS model customers, staff, and fitting rooms were all active objects (agents) in order to allow communication among them through message passing. In both simulation models, we mimicked the real reactive and proactive behaviour of a sale staff towards the customer.

The customer arrival rate as we observed in the real system during a typical day is as shown in Figure 4 below.

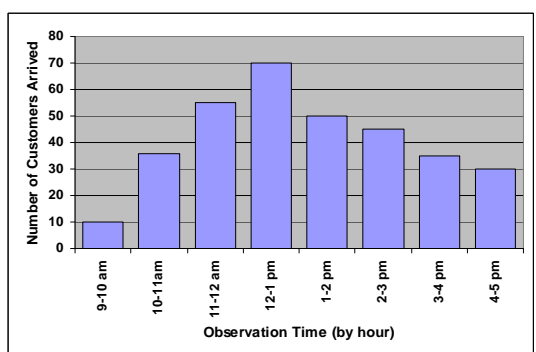

**Figure 4** Distribution of customer arrival in the real system on a typical day

In our simulation models, we modelled the arrival process using an exponential distribution with an hourly changing arrival rate in accordance with Figure 4. The simulation models were terminated after a standard business day (8 hours), mimicking the operation of the real department store. We conducted 100 replications for each set of parameters. Both simulation models used similar model input parameter values. Therefore, if any differences in the model outputs were noticed, they would be only due to the differences between the two modelling techniques.

### 4.2 Reactive and Proactive Behaviour Setup

Reactive and proactive behaviours for both simulation models were set up similarly. We pointed one staff member that performed all three reactive jobs mentioned in Section 2, namely job 1 (counting garments on entry), job 2 (providing help) and job 3 (counting garments on exit). The staff served the customers by first come first serve approach. There were few cases where we considered proactive behaviour (staff changes their service times from normal to fast) when there were customers queuing while fitting room cubicles were available or to get served by the staff. In order to speed up the servicing time, we reduced the normal service time by 20%. We implemented this proactive behaviour using the procedures shown in Figure 5. The decision making was done based on a set of selection rules (using decision tree solutions) and probabilistic distribution. Each block in Figure 5 represents the event as shown in Figure 6.

Condition in the fitting room and number of waiting customers in the three queues were checked continuously via probability distribution. When the condition was met, the service time was speeded up automatically. After some delay performed by the probability distribution, the new service time was changed to the existing service time.

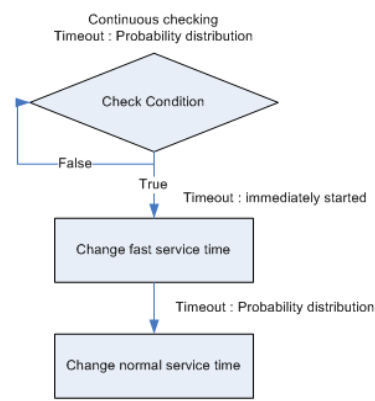

**Figure 5** Proactive decision making

```
Event Check Condition
For < all fitting room cubicles >
  If < fitting room cubicles is busy =
      false && customer waiting in
      entry queue >= number waiting >
      start event change service time
      without delay;
  else
  If < customer waiting in return queue
      >= number waiting >
      start event change service time
      without delay;
  else
  If < customer waiting in help queue >=
      number waiting >
      start event change service time
      without delay;

Event Change Service Time
For < staff >
      existing service time = new
      service time;
      count the service time changes;
      start event change to existing
      service time by delay
      (probability distribution);

Event Change To Existing Service Time
For < staff >
      existing service time = existing
      service time;
```

**Figure 6** Proactive Behaviour Pseudo Code

### 4.3 Experiment 1: Sensitivity Analysis

We validated both our simulation models by conducting a sensitivity analysis where the customer arrival rates were systematically varied (we increased it by 30% each time) and we observed how this affected our system performance measures (customer waiting times from three queues, staff utilisation, number of service time changes, cubicle utilisation, number of customer served and number of customer not been served).

#### 4.3.1 Sensitivity Analysis
Results for the sensitivity analysis are shown in Table 1 and Figure 7. The results in both figure and table illustrate that there were some similar patterns for all performance measures. Both simulation models demonstrated an increment for all performance measures when the customer arrival rate was increased. We found, there were no differences in the results of number of service time change in Figure 7 between DES and DES/ABS models. However, from the bar chart below shows that the DES/ABS model served more customers compared to the DES model. This result also means that the DES/ABS produces lower customer waiting times, higher staff serving utilisation, lower number of customer not served and higher cabin utilisation compared to the DES model.

| Simulation Models | Number of cubicles | Performance measures | | No of customer arrival (per day) | | | | |
|---|---|---|---|---|---|---|---|---|
| | | | | 1 | 2 | 3 | 4 | 5 |
| DES | 8 | Waiting time | Mean | 1.32 | 7.68 | 11.90 | 15.10 | 16.14 |
| | | | SD | 0.92 | 1.56 | 1.37 | 0.96 | 0.44 |
| | | Staff utilisation | Mean | 44% | 52% | 55% | 57% | 59% |
| | | | SD | 6.30 | 6.28 | 6.37 | 7.08 | 6.31 |
| | | Cubicle utilisation | Mean | 66% | 80% | 88% | 91% | 93% |
| | | | SD | 4.05 | 3.28 | 1.88 | 1.19 | 1.24 |
| | | Number not served | Mean | 5 | 33 | 111 | 249 | 446 |
| | | | SD | 4.37 | 14.12 | 21.35 | 26.28 | 28.82 |
| | | Number served | Mean | 312 | 386 | 430 | 445 | 451 |
| | | | SD | 18.54 | 17.00 | 12.72 | 13.14 | 12.42 |
| | | Service time changes | Mean | 26 | 57 | 80 | 86 | 93 |
| | | | SD | 14.41 | 17.53 | 22.55 | 24.02 | 20.80 |
| DES/ABS | 8 | Waiting time | Mean | 1.21 | 5.44 | 9.58 | 12.97 | 13.96 |
| | | | SD | 1.15 | 2.83 | 2.89 | 2.55 | 1.98 |
| | | Staff utilisation | Mean | 46% | 59% | 62% | 68% | 69% |
| | | | SD | 7.88 | 7.43 | 8.22 | 11.00 | 10.54 |
| | | Cubicle Utilisation | Mean | 66% | 83% | 91% | 97% | 98% |
| | | | SD | 10.34 | 6.75 | 5.29 | 3.28 | 1.62 |
| | | Number not served | Mean | 9 | 26 | 93 | 229 | 412 |
| | | | SD | 15.79 | 24.48 | 43.26 | 63.92 | 79.71 |
| | | Number served | Mean | 300 | 397 | 456 | 486 | 514 |
| | | | SD | 49.70 | 23.62 | 39.18 | 58.46 | 74.57 |
| | | Service time changes | Mean | 24 | 56 | 85 | 88 | 95 |
| | | | SD | 15.54 | 19.18 | 27.73 | 38.32 | 57.86 |

**Table 1** Result of Sensitivity Analysis Experiment

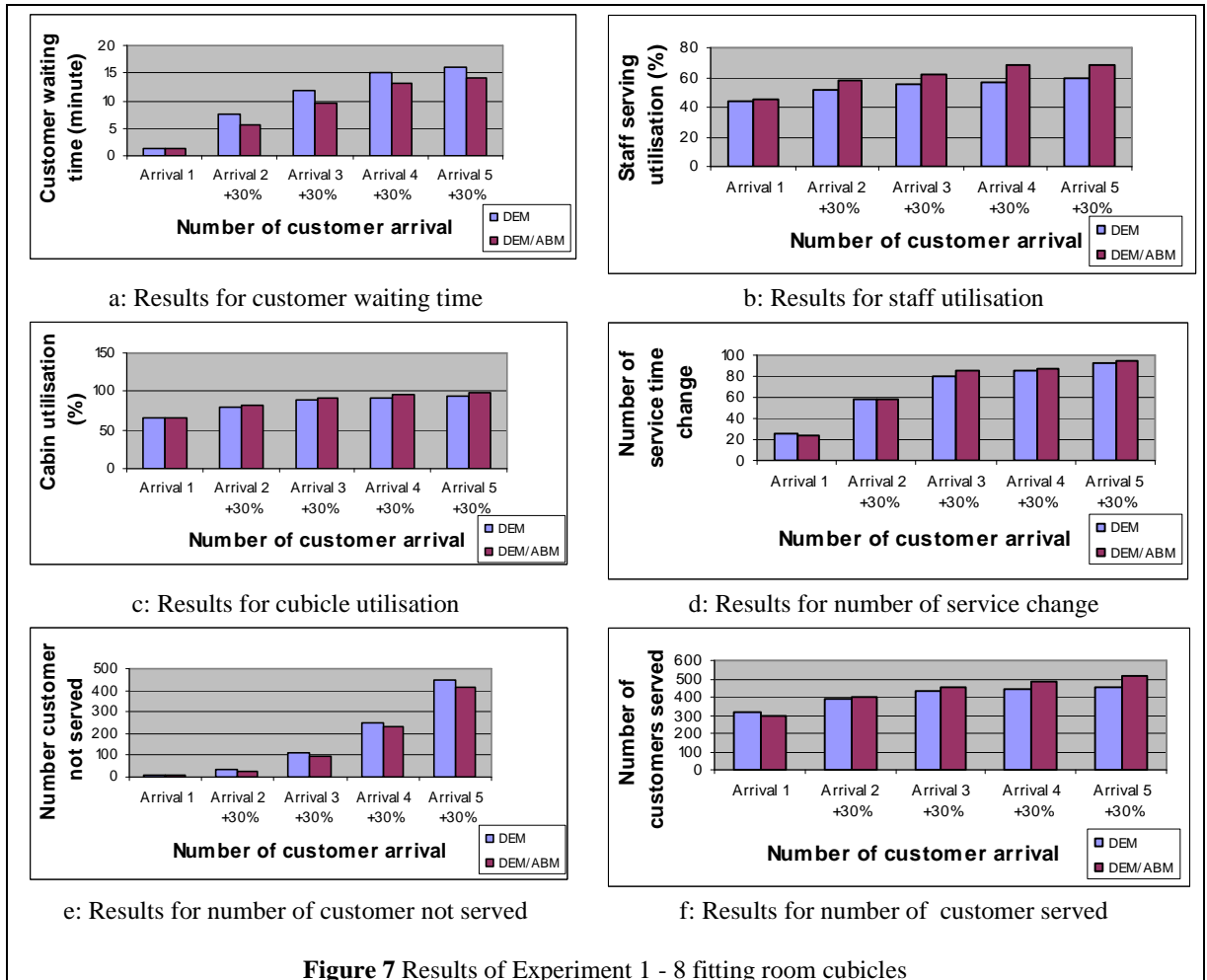

**Figure 7** Results of Experiment 1 - 8 fitting room cubicles

Hence, we conclude that the sensitivity analysis has proven that staff's proactive behaviour has significant impact on both simulation models. However, the observation has also discovered that the impact on the output performance of the DES/ABS model is slightly higher than DES model. We will investigate this further in the future.

### 4.4 Comparing the Impact of Reactive and Mixed Reactive and Proactive Behaviour

In a previous experiment (Majid et al, 2009), namely Experiment A, we modelled exclusively the reactive behaviour of sales staff towards a customer using DES and DES/ABS. As to improve the model, the first experiment in this paper (see Section 4.3), namely Experiment B, we modelled a mix of reactive and proactive behaviour of sales staff towards a customer. This time, we aimed to make statistical comparison of the difference between modelling reactive behaviour and the mixed of reactive and proactive behaviours involved in our simulation models. This comparison is vital to determine the similarities and dissimilarities of both models in the output performance.

We considered similar scenarios for both experiments: We had one staff member that performed three tasks ((1) counting garments on entry, (2) providing help, and (3) counting garments on exit), we set a fixed number of customers arriving per day (300) and we fixed number of fitting room cubicles (8). We added another proactive feature which was the demand driven change in service time. We selected customer waiting time and staff utilisation as performance measures for our statistical comparison. Our hypotheses are as below:

- $Ho_1$ = The average customer waiting times resulted from our DES model are not

- significantly different in Experiment A and B.
- $Ho_2$ = The average customer waiting times resulted from our DES/ABS model are not significantly different in Experiment A and B.
- $Ho_3$ = The staff utilisation values resulted from our DES model are not significantly different in Experiment A and B.
- $Ho_4$ = The staff utilisation values resulted from our DES/ABS model are not significantly different in Experiment A and B.

As our data was not normally distributed, the mean and median values would be different. Hence, in order to compare the median values we chose the non parametric Mann-Whitney statistical test. This method helped to confirm or disconfirm the above hypotheses. For performing the Mann-Whitney test, we applied the open source statistical software package R (The R Foundation for Statistic Computing, 2009). The median of both performance measures in DES for both reactive and proactive experiment were calculated for this test. The similar procedure was done for the DES/ABS. We chose 0.05 as our significance level. A test result (p-value) higher than 0.05 would allow us to accept a null hypothesis; otherwise we would have to reject it. Testing our DES model results on customer waiting times and staff utilisation for Experiment A against Experiment B revealed a p-value of 0.1608 and 0.000 respectively. Since the DES p-value for waiting times was greater than our chosen significance level (0.05) we failed to reject our $Ho_1$ hypothesis. In contrast, we have to reject $Ho_3$ hypothesis as the DES p-value is lower than our chosen significance level (0.05).

Testing our DES/ABS model results on customer waiting times and staff utilisation for Experiment A against Experiment B reveals a p-value of 0.06 and 0.000 respectively. In similar case with DES model results, we failed to reject our $Ho_2$ hypothesis for waiting times as the DES/ABS p-value was higher than our chosen level of significance (0.05). Meanwhile, for staff utilisation, the DES/ABS p-value was lower than our chosen significance level (0.05) which $Ho_4$ hypothesis had to be rejected.

On the basis of our statistical results on the measures of central tendency, it can be concluded that there was significant difference between the average customer waiting times and staff utilisation in Experiment A and Experiment B for both DES and DES/ABS models. This implies that proactive behaviour can help a staff to reduce the servicing time and providing a bigger impact on reducing the staff utilisation. However, being proactive by reducing the servicing time only shows limited impact on customer waiting times in both simulation models. This is because the number of fitting room cubicles is a further bottleneck. We will investigate this issue further by adding another proactive behaviour which; the staff asks another staff for help when she meets certain condition. We expect that this time we can see differences in customer waiting times.

As an overall conclusion of our reactive compared to mixed reactive and proactive experiments, we found that there are no differences between the DES and DES/ABS models' output performance when modelling similar human behaviour.

## 5. CONCLUSION AND FUTURE WORK

This paper has presented simulation as a tool to investigate the impact of human reactive and proactive service behaviour in the retail sector. As a matter of fact, we were more interested in finding out the benefits of implementing one behaviour or the other. Our investigations focused on determining the advantages and disadvantages of implementing the reactive and proactive service behaviour in a simulation model.

We dealt with the reactive service behaviour in an earlier paper (Majid et al 2009) and in this paper we focused on the proactive service behaviour. In order to understand the impact of implementing the proactive service behaviour to the real system, we modelled the behaviour for DES and DES/ABS. Proactive service behaviour relates to a staff making an autonomous decision in handling an uncontrolled situation in the fitting room (in our case reducing the service times and consequently reduce customer waiting times). In order to gain a valid proactive model, we conducted the sensitivity analysis where we firstly varied the number of customer arrivals and secondly the number of fitting cubicles. The analysis discovered that the proactive staff behaviour has affected both; the performance measures of the DES model as well as the performance measures of the DES/ABS model. However, the study found that the impact on the DES/ABS model was much noticeable. We then compared our simulation results from the sensitivity analysis (proactive behaviour) with results from an experiment (reactive behaviour) that we conducted in previous paper (Majid et al 2009). It has been discovered that there was not significant difference in the customer waiting times output resulted from DES and DES/ABS model when considering either reactive or mixed reactive and proactive service behaviour. Nonetheless, the difference was significant in the performance output of staff utilisation in the DES and DES/ABS models. This demonstrates that both simulations approaches show similarities in modelling and simulating the similar human behaviour.

In future, we will investigate further about the differences of results that we found in the sensitivity analysis experiment between DES and DES/ABS models. In addition, we would like to include other forms of proactive staff behaviour in our simulation models. We are also planning to start with our second case study soon; this time in public sector, to examine if we can generalise our findings regarding the usefulness of combined DES/ABS in investigating the behaviour of human centric complex systems.

**AUTHORS BIOGRAPHIES**

**MAZLINA ABDUL MAJID** is a PhD student in the School of Computer Science, University of Nottingham. Her interests are in discrete event simulation and agent based simulation. Her email is <mva@cs.nott.ac.uk>.

**PEER –OLAF SIEBERS** is a Research Fellow in the School of Computer Science, University of Nottingham. His main interests include agent based simulation and human complex adaptive system. His email is <pos@cs.nott.ac.uk>.

**UWE AICKELIN** is a Professor in the School of Computer Science, University of Nottingham. His interests include agent based simulation, heuristics optimisation, artificial immune system. His email is <uxa@cs.nott.ac.uk>.